# Building a Multimodal Dataset of Academic Paper for Keyword Extraction

**Zhang, Jingyu** Department of Information Management, Nanjing University of Science and Technology, China | zzjy@njust.edu.cn

**Yan, Xinyi** Department of Information Management, Nanjing University of Science and Technology, China | yanxinyi@njust.edu.cn

**Xiang, Yi** Department of Information Management, Nanjing University of Science and Technology, China | xiangyi@njust.edu.cn

**Zhang, Yingyi** Department of Archives and E-government, Soochow University, China | yyzhang9@suda.edu.cn

**Zhang, Chengzhi*** Department of Information Management, Nanjing University of Science and Technology, China | zhangcz@njust.edu.cn

## ABSTRACT

Up to this point, keyword extraction task typically relies solely on textual data. Neglecting visual details and audio features from image and audio modalities leads to deficiencies in information richness and overlooks potential correlations, thereby constraining the model's ability to learn representations of the data and the accuracy of model predictions. Furthermore, the currently available multimodal datasets for keyword extraction task are particularly scarce, further hindering the progress of research on multimodal keyword extraction task. Therefore, this study constructs a multimodal dataset of academic paper consisting of 1000 samples, with each sample containing paper text, images, audios and keywords. Based on unsupervised and supervised methods of keyword extraction, experiments are conducted using textual data from papers, as well as text extracted from images and audio. The aim is to investigate the differences in performance in keyword extraction task with respect to different modal information and the fusion of multimodal information. The experimental results indicate that text from different modalities exhibits distinct characteristics in the model. The concatenation of paper text, image text and audio text can effectively enhance the keyword extraction performance of academic papers.

## INTRODUCTION

The rapid development of information technology has led to the exponential growth of scientific and technological information. This information is presented and disseminated in various modalities such as text, image, video and audio, constituting the diversity and multidimensional characteristics of the information field. The abundance of multimodal data has led to issues such as data explosion and information overload, resulting in negative impacts such as resource dispersion, quality degradation and decision-making difficulties. Extracting key information can help filter and quickly understand important content behind the data, thereby effectively addressing these problems. Therefore, extracting key information effectively from vast multimodal data remains a crucial challenge to be addressed.

Keywords accurately describe the core thematic information in textual modalities and typically comprise individual words or short phrase that succinctly summarize the content. Currently, most texts (e.g. academic papers, texts in Twitter and Weibo) have deficiencies in additional annotated keywords, resulting in problems such as information redundancy and disorganized information management. However, keywords possess advantages such as aiding readers in quickly grasping the main topics and core content of information, facilitating information organization (Meng et al., 2017) and so on. Therefore, it is necessary to extract keywords from data. So far, keyword extraction task typically relies on textual data. However, in practice, data is often presented in a multimodal manner. This poses challenges to traditional text-based keyword extraction methods. Moreover, the lack of multimodal datasets for keyword extraction task further impedes the progress of research in multimodal keyword extraction. Therefore, this paper constructs a multimodal dataset of academic paper for keyword extraction. It includes conference papers, presentation slides, speaker audio and keywords.

The dataset comprises textual, image and audio modalities, each conveying rich information from different perspectives. To fully leverage the information from different modalities and exploit the correlations between them, we need a multimodal keyword extraction method with powerful data representation learning capabilities. For example, integrating information from the image modality into the keyword extraction task (Wang et al., 2020), along with combining web page text and visual features, using a supervised approach for training and prediction, experimental results have shown that incorporating visual features supposed to be enhance the effectiveness of keyword extraction. Aligning video frames and audio from scientific video lectures (Hassani et al., 2022), merging image and speech text content, and comprehensively considering both local and global features of candidate phrases, the results indicate that combining text from both modalities can enhance the effectiveness of keyword extraction. Based on unsupervised and supervised methods of keyword extraction, this paper conducts experiments to explore the differences in performance of keyword extraction task when utilizing text information from different modalities and integrating multimodal text information.

* Corresponding author: Chengzhi Zhang (zhangcz@njust.edu.cn).

Up to this point, there is a shortage of multimodal datasets specifically designed for keyword extraction task. This paper first constructs a multimodal dataset of academic papers. Both unsupervised and supervised keyword extraction experiments are conducted using textual data from papers, as well as text extracted from images and audio, to validate the practical effectiveness of the dataset.

The main contributions of this study can be summarized as follows:

Firstly, the focus is on building a multimodal dataset of academic papers focuses on keyword extraction task. Due to the specialized nature of academic papers and the presence of keywords within them, the dataset exhibits high quality and reliability. Therefore, multimodal academic papers are utilized to construct the dataset. This work fills the gap of the absence of datasets tailored specifically for keyword extraction task and provides new data resources for multimodal academic research. It also helps to advance keyword extraction research towards the utilization of a broader range of modal data, enhancing algorithm accuracy and robustness.

Secondly, the concatenation of paper text, image text and audio text can effectively enhance the keyword extraction performance. This indicates that information from different modalities can complement and enhance each other. By concatenating text information from multiple modalities, a more comprehensive capture of document semantics can be achieved, thereby improving the accuracy of keyword extraction. By integrating information from different modalities, new methods and perspectives can be provided for multimodal keyword extraction research.

## RELATED WORK

This paper aims to construct a multimodal dataset for keyword extraction task based on academic texts and further investigate the performance impact of multimodal data fusion strategies on keyword extraction. Therefore, the research covered in this paper includes studies on both text-based keyword extraction and multimodal keyword extraction.

### Textual Keyword Extraction

The research methods currently used for text-based keyword extraction can be divided into unsupervised methods and supervised methods (Beliga, 2014; Papagiannopoulou & Tsoumakas, 2020). Supervised keyword extraction methods utilize annotated data for model training to learn the patterns of keyword extraction. It achieves accurate identification and extraction of keywords from unlabeled text by establishing a model that maps the relationship between input text and keyword labels. The traditional machine learning model KEA (Witten et al., 1999) is a classic method for keyword extraction. It utilizes feature engineering to form feature vectors using TF-IDF values, word frequency and word position in the document. Subsequently, it employs the Naive Bayes classification algorithm for word type classification. The sequence labeling method which called Ranking SVM (Jiang et al., 2009), transforms candidate phrases into feature vector representations and then employs a support vector machine model to classify them into two categories: keywords and non-keywords. A deep learning model (Tu et al., 2016) is introduced with a coverage mechanism to determine whether all the keywords for all topics in the document have been extracted and whether there are any semantic duplicates among the keywords. BiLSTM (Basaldella et al., 2018) considers both forward and backward propagation of text information simultaneously to enable better recognition of structural information. Studies have demonstrated that the synergy of a BiLSTM layer and a CRF (Conditional Random Field) layer for sequence labeling significantly enhances performance outcomes. The JointKPE model for open-domain keyword extraction, built upon the RoBERTa model (Sun et al., 2021), supposed to be simultaneously capture both local phrases and global information. The Weibo keyword extraction method (Zhang et al., 2020) utilizes rich text context from dialogues and designs a neural network framework consisting of a dialogue context encoder and a keyword tagger. These modules are jointly trained to optimize the keyword extraction process, employing hierarchical structures and character-level representations to process Weibo text. Large language models typically have billions or more parameters, which enables them to learn increasingly complex linguistic structures and exhibits robust of self-learning and generalization capabilities. A method is proposed for generating keywords using Galactica, a pre-trained large language model published by Meta (Lee et al., 2023), which performs better than previous studies.

Unsupervised methods extract significant words from text as candidate keywords autonomously, without reliance on manually annotated corpora. These methods rank the candidate words according to predefined rules and then select the top-ranked candidates as keywords (Hasan & Ng, 2010, Merrouni et al., 2016). Statistical-based approaches derive keywords by analyzing textual statistical information. The Term Frequency-Inverse Document Frequency (TF-IDF) (Frank et al., 1999) method gauges words importance in a document by calculating their term frequency and inverse document frequency. YAKE (Campos et al., 2018) integrates the positional information of keywords, and employs innovative statistical metrics to capture the contextual information of candidate words. KCRank (Won et al., 2019) is a statistical method for extracting keywords at document level, combining simple heuristic rules, which can compete with state-of-the-art systems. Graph-based methods involve constructing a network of candidate words and identifying pivotal nodes to determine keywords. TextRank (Mihalcea & Tarau, 2004) treats words in the document as graph nodes, with word co-occurrence forming the graph edges. Through iterative calculation and node values updates, words with higher values are designated as keywords for the text. SGRank (Danesh et al., 2015) and PositionRank (Florescu & Caragea, 2017) consider a more comprehensive range of statistical indicators. CiteTextRank (Gollapalli & Caragea, 2014) employs citation networks to expand contextual information and simultaneously extracts keywords based on topic modeling. TopicPageRank (Ding et al., 2011) makes use of topic modeling to discern document topics and their significance, while also identifying relevant words, assigning higher initial values to words associated with critical topics. SIFRank (Sun et al., 2020) is a new baseline for unsupervised keyword extraction based on pre-trained language model.

Deep learning models outstrip unsupervised methods in keyword extraction task. Nevertheless, their efficacy hinges on the scale of data and relies on annotated data, which can prove costly and prone to consistency issues in large-scale annotation endeavors. Conversely, Unsupervised methods sidestep the need for extensive annotated data and instead harness intrinsic statistical or structural information within the text to construct models, endowing them with versatility and broad applicability. Nonetheless, unsupervised methods often grapple with capturing profound semantic nuances within the text, potentially overly accentuating local structures while neglecting global contextual information. Tuning parameters poses a challenge, hindering the attainment of optimal performance.

### Multi-modal Keyword Extraction

In video tagging research, some methods rely on extracting text from multimedia sources and then extracting keywords from the textual content (Awad et al., 2017), while others consider annotations, video frames and audio signals. In the research on retrieving scientific lecture videos (Radha, 2016), information is extracted from video frames and segmented. Optical Character Recognition (OCR) and Automatic Speech Recognition (ASR) technologies are used to extract textual information from both video frames and audio. Specifically, the Sphinx speech recognition system is employed to extract text from audio, whilst text extraction from frames is accomplished through image matching. The video captioning algorithm (Nabati & Behrad, 2020) employs deep neural networks and image processing to generate multi-sentence descriptions based on the content of video clips. It employs beam search algorithm and content structure filtering to remove erroneous sentences. Combining semantic attention and temporal attention for video captioning, a network incorporating GRU-based semantic temporal attention is proposed (Jin et al., 2019). This network applies high-level visual concepts to construct semantic temporal attention. The video captioning algorithm (Chen et al., 2019) makes use of multi-modal features such as images, motion, auditory and speech descriptions. During this process, it trained a topic prediction model for decoding and generating subtitles, and then utilized this model for predictions. Within each video topic, it employs an LSTM network and combines these topic-specific LSTM networks to construct a hybrid topic LSTM network. The corr-LDA (Iyer et al., 2016) method leverages audio content for content-based video indexing tasks. It assigns key phrases to videos using a dictionary constructed from training texts. Using human reading time to extract keywords from Weibo (Zhang & Zhang, 2021), the study extracts reading time data through eye-tracking and integrates it into the keyword extraction model. Two neural network models are proposed for this purpose. A method that combines electroencephalogram (EEG) signals and eye-tracking signals (Yan et al., 2024) to improve automatic keyword extraction effectiveness. Key features are selected from cognitive signals generated during human reading, and through multiple tests, it has been found that EEG signals can significantly enhance keyword extraction performance.

Most existing research primarily focuses on extracting keywords from text, which has made progress. However, it overlooks information from other modalities such as image and audio which leads to a lack of richness in information and neglect of potential correlations, thereby limiting the model's ability to learn data representations accurately and predict outcomes effectively. Nonetheless, research on multimodal keyword extraction, which integrates information from various modalities such as eye-tracking signals, EEG signals, motion and others, can lead to a more comprehensive learning of data representations, thereby enhancing the accuracy of keyword extraction. Yet, there is limited research on multimodal keyword extraction, possibly due to the complexity of feature extraction and model design for multimodal data, as well as challenges in data acquisition and annotation.

## CONSTRUCTION OF MULTIMODAL DATASETS

This paper aims to construct a multimodal dataset of academic paper, uncovering the differences in keyword extraction task among different modal text information, and exploring the impact of text fusion of multimodal data on keyword extraction performance. The research framework is illustrated in Figure 1. The first part involves constructing a multimodal dataset of academic paper, employing techniques such as web scraping, image processing and text mining to build a multimodal dataset that based on text, image and audio. The second part focuses on building unsupervised and supervised keyword extraction models. The third part includes conducting experiments using paper text, audio text, image text and the combinations of the three text modalities on the established models.

### Selection of Data Source

Existing keyword extraction datasets are predominantly text-based, lacking multimodal datasets to support keyword extraction task. Hence, this study builds a dataset comprising image, audio and text modalities. Using academic conferences as a data source that offers several advantages. Firstly, they typically encompass multiple modalities of information, including textual content from conference papers, presentation slides, and audio recordings of speakers. Secondly, they possess expertise and depth, as academic conference content often pertains to specialized knowledge and research findings within specific fields, providing authoritative information within those domains. Considering high-definition videos, high-quality audio and transcriptions of audio texts can sure the quality of the dataset, this study selects VideoLectures (*https://videolectures.net*) and the SPIE (International Society for Optics and Photonics) Digital Library (*https://www.spiedigitallibrary.org*) as data platforms. In this section, data is manually selected from these sources. VideoLectures is an English online platform offering a wide range of free educational video resources, including lectures, speeches, and courses across various academic disciplines. The SPIE Digital Library, provided by the International Society for Optics and Photonics, is an online platform hosting various academic journals and conference proceedings published by SPIE. It covers the latest research findings and cutting-edge technologies in fields such as optical engineering, photonics, sensor technology, medical imaging, among others.

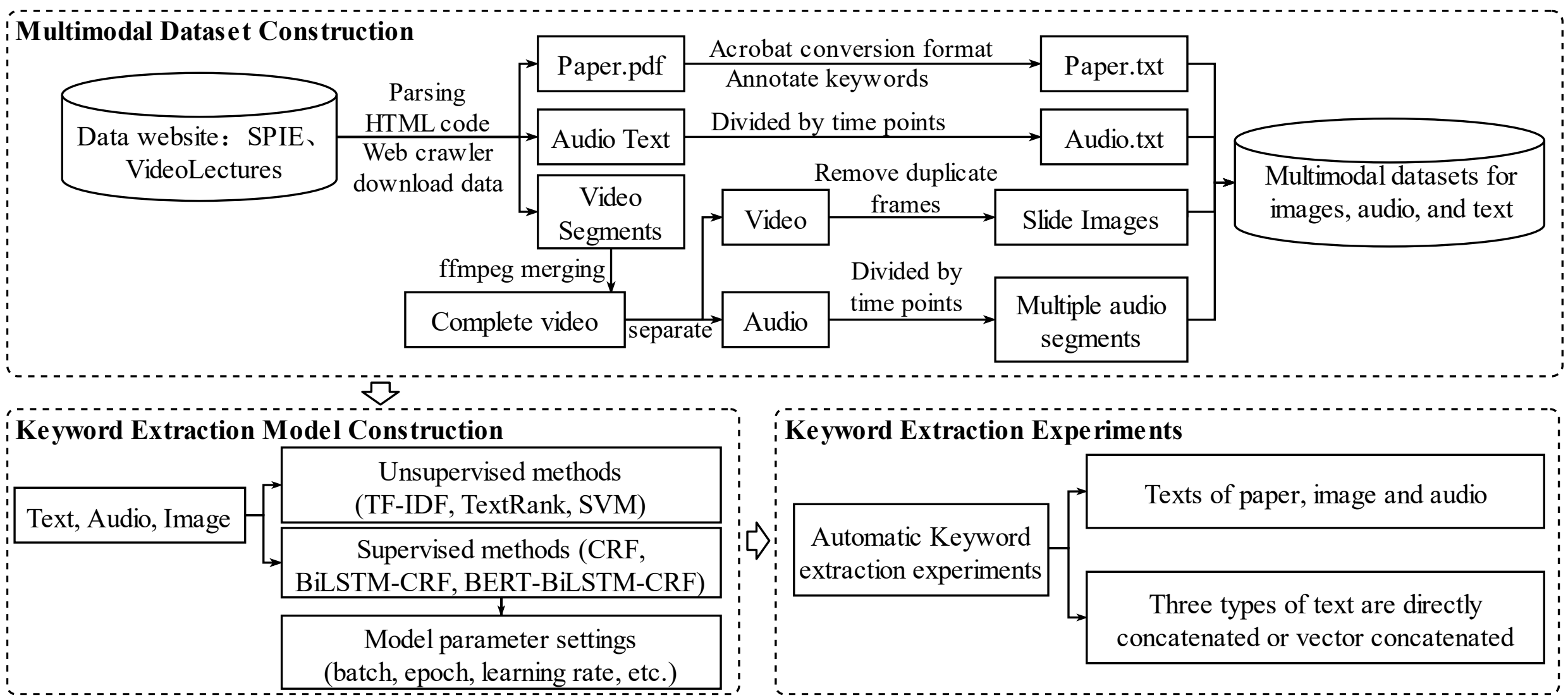

**Figure 1. Research Framework**

### Data Collection and Preprocessing

With the use of web scraping tools, academic websites can be accessed to extract videos, papers and audio texts. Videos on these sites are accessible only via one of two streaming protocols: HLS (HTTP Live Streaming) or DASH (Dynamic Adaptive Streaming over HTTP). The specific details about video loading are contained within m3u8 and mpd files. By parsing the files' content, both video and audio segments can obtain. Then, employ the ffmpeg (Fast Forward MPEG, *https://ffmpeg.org*) library to combine them into complete videos for further analysis. Audio transcripts need to be dynamically sourced from the HTML source code. Papers are accessible via direct links.

Academic conference videos need to be processed into image and audio formats. Firstly, use a video processing library like OpenCV to separate the audio signal from the video. Secondly, the video is composed of a series of video frames, and the content in academic conference videos usually consists of slides, with each slide being displayed for a certain amount of time. Therefore, there are duplicate frames in the video. Using image processing techniques such as structural similarity algorithms to compare the similarity of video frames can effectively remove duplicate frames. Then, save the first occurrence of each video frame. Animations in the slides may affect the accuracy of extraction, so manual verification is needed to obtain complete slides. During the alignment process of images and audio, the audio segments are accurately segmented based on the playback time of different video frames. This is achieved using audio processing libraries such as LibROSA (*https://librosa.org*). The goal is to align the image segments of slides with the corresponding audio segments. Similarly, alignment between the speech text and slide images can be achieved based on the time information in the speech text, thereby establishing the correlation between the image and audio modalities. In the preprocessing stage of the paper, Acrobat (*https://acrobat.adobe.com*) is used to convert PDF-format papers into TXT-format, and annotations are added for keywords and other content. Finally, data cleaning is performed using a heuristic rule-based approach, including removing garbled characters, line breaks, etc. An example of the dataset is illustrated in Figure 2.

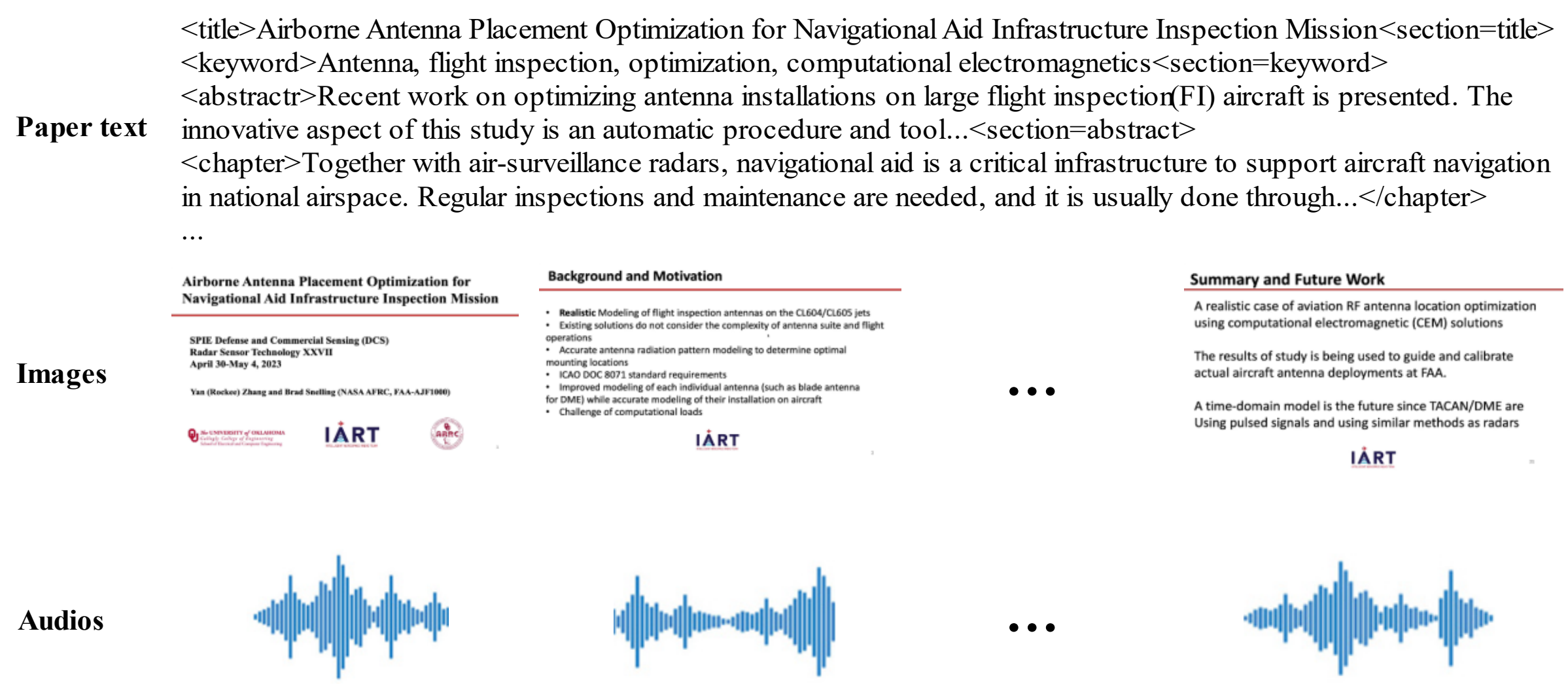

***Note***: Full content of sample paper can be accessed via: *https://doi.org/10.1117/12.2663414*

**Figure 2. Data Sample**

## Descriptive Statistics of the Multimodal Dataset

The multimodal dataset of academic paper comprises 1, 000 samples, each containing paper text, text keywords, slide images and audio. Table 1 describes the maximum, minimum and average lengths of each type of data. The maximum length of paper text is 9313 words, the minimum length is 1084 words, and the average length is 4051 words. The maximum number of keywords is 11, the minimum is 2, and the average number of keywords is 5.4. The number of slides refers to its' pages, with a maximum of 41 pages, a minimum of 9 pages, and an average of 23.5 pages. The complete audio time ranges from a maximum of 36min32s to a minimum of 9min12s, with an average duration of 18min17s.

| Data Type | Min Length | Max length | Average Length |
|---|---|---|---|
| Paper Text | 1084 words | 9313 words | 4051 words |
| Keywords | 2 | 11 | 5.4 |
| Slides | 9 pages | 41 pages | 23.5 pages |
| Audio | 9min12s | 36min32s | 18min17s |

**Table 1. Statistics on the Length of Various Types of Data**

Figure 3 reflects the distribution of lengths across different types of data.

Figure 3(a) illustrates the distribution of word count in the dataset's papers. By observing, it can be inferred that there is a higher frequency of papers with word counts ranging from 2000 to 5500, while there is a lower frequency of papers with word counts exceeding 7500 or falling below 2000. This suggests that the majority of academic papers tend to have lengths concentrated within a moderate range. Such moderate-length texts often contain richer semantic information, thereby offering more contextual cues, which in turn facilitates the accurate extraction of keywords by models.

Figure 3(b) describes the distribution of keyword counts in the papers. Through observation, it is evident that there is a higher frequency of papers with a keyword count ranging from 3 to 8, while there is a lower frequency of papers with keyword counts exceeding 8 or falling below 2. This indicates that the majority of academic papers tend to have a moderate number of keywords. Such moderate number of keywords allows for a concise yet comprehensive representation of the paper's themes and content, contributing to the training and optimization of keyword extraction models.

Figure 3(c) shows the distribution of slide page numbers. Through observation, it is evident that there is a higher frequency of slides with images count ranging from 13 to 25, while there is a lower frequency of slides with images count exceeding 31 or falling below 13. This indicates that the majority of slides tend to have a moderate number of images. Such a moderate number of images facilitates effective visual support for the presentation content, striking a balance between information density and clarity.

Figure 3(d) depicts the distribution of audio durations. The observation reveals that there is a higher frequency of audio recordings with durations ranging from 12 to 21 minutes, while there is a lower frequency of recordings exceeding 27 minutes or falling below 12 minutes. This indicates that the majority of audio recordings tend to have durations concentrated within a moderate range. Such moderate durations facilitate effective communication of content, striking a balance between avoiding excessive length that could lead to information scarcity and preventing brevity that might compromise the integrity of information.

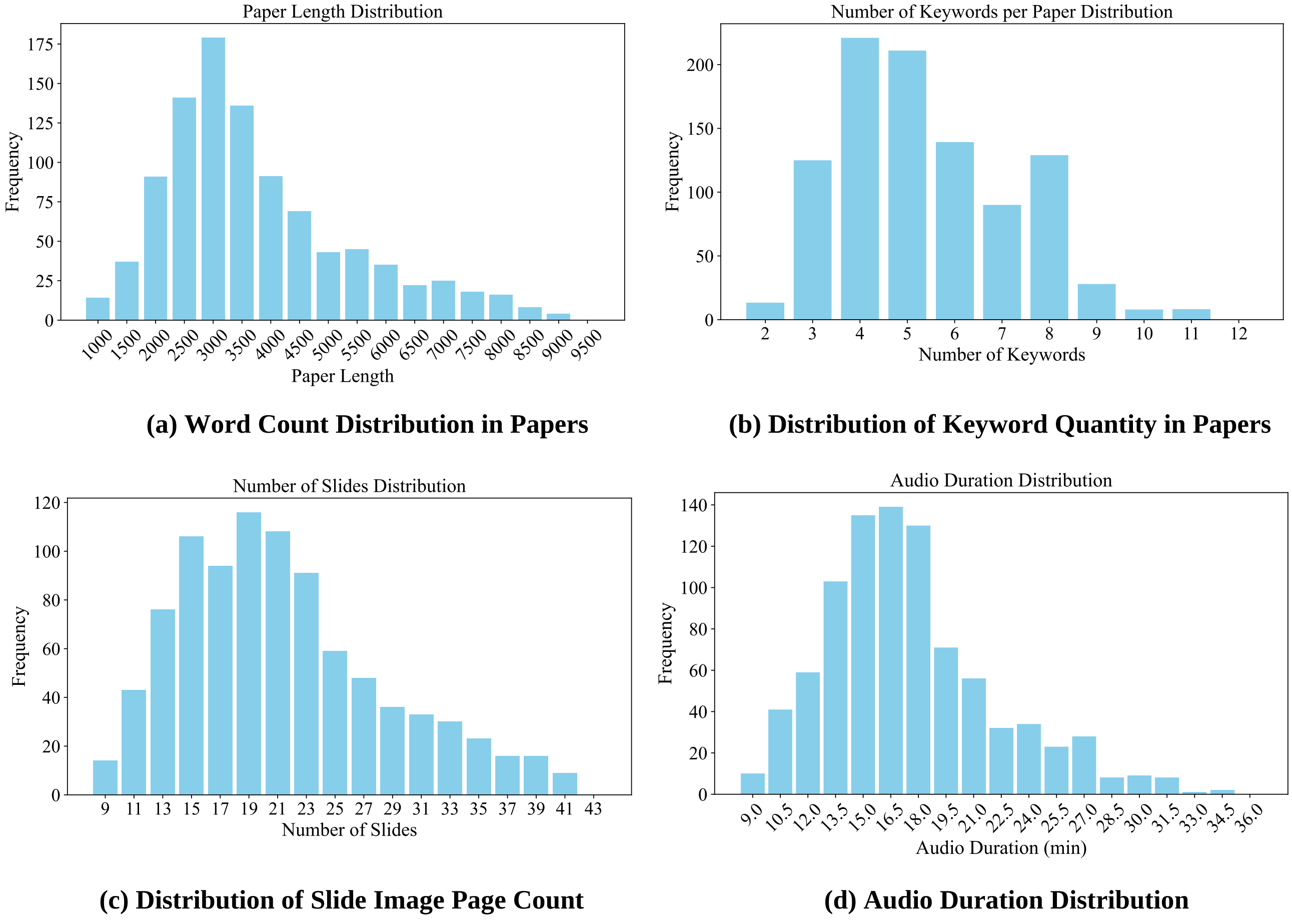

**(a) Word Count Distribution in Papers**

**(b) Distribution of Keyword Quantity in Papers**

**(c) Distribution of Slide Image Page Count**

**(d) Audio Duration Distribution**

**Figure 3. Overview of the Distribution of Lengths for Each Data Type**

## KEYWORD EXTRACTION BASED ON MULTIMODEL DATASET

In this section, experiments are conducted using the constructed multimodal dataset of academic paper to explore the practical effectiveness of the dataset in keyword extraction based on both unsupervised and supervised methods.

### Keyword Extraction Models and Evaluation Methods

This section introduces the unsupervised and supervised keyword extraction methods used in the experiments, and adopts the method of exact matching to assess the effectiveness of keyword extraction models.

#### *Keyword Extraction Models*

The unsupervised keyword extraction methods include TF-IDF, TextRank and SVM (Zhang et al., 2006). The supervised methods consist of CRF (Zhang, 2008), BiLSTM-CRF (Alzaidy et al., 2019) and BERT-BiLSTM-CRF (Dai et al., 2019). Among them, BiLSTM-CRF is a sequence labeling model commonly used for tasks like keyword extraction and named entity recognition. Figure 4 illustrates the model architecture of BiLSTM-CRF. It mainly consists of the input layer, embedding layer, BiLSTM layer, linear layer, CRF layer and output layer. In the input layer, the model receives the positional indices $\{w_1, w_2, \ldots, w_n\}$ of each word in the text sequence. The embedding layer maps the positional indices of each word to corresponding word vectors. The BiLSTM layer captures the semantic relationships between words, resulting in semantic vectors for each word $\{h_1, h_2, \ldots, h_n\}$. The linear layer converts the semantic vectors of each word into probability scores for four types of tags ([PAD], B, I, O), which are then used as the unnormalized emission probabilities in the CRF layer. The CRF layer learns the dependency information between labels to ensure the rationality of predicted labels. The output layer generates the classification labels for each word, where [PAD] represents the category for padded words, and the meanings of the B, I and O labels are consistent with those used in the CRF model. In BERT-BiLSTM-CRF, the BERT pre-trained language model replaces the word embedding layer to obtain word vectors for each word, while the other layers remain unchanged. In the experiments, to reduce memory consumption, the input text content is segmented into multiple text fragments. Each fragment has a length of 128 words and is slid with a stride of 64 words. Subsequently, each text fragment is annotated based on the provided keywords by the authors.

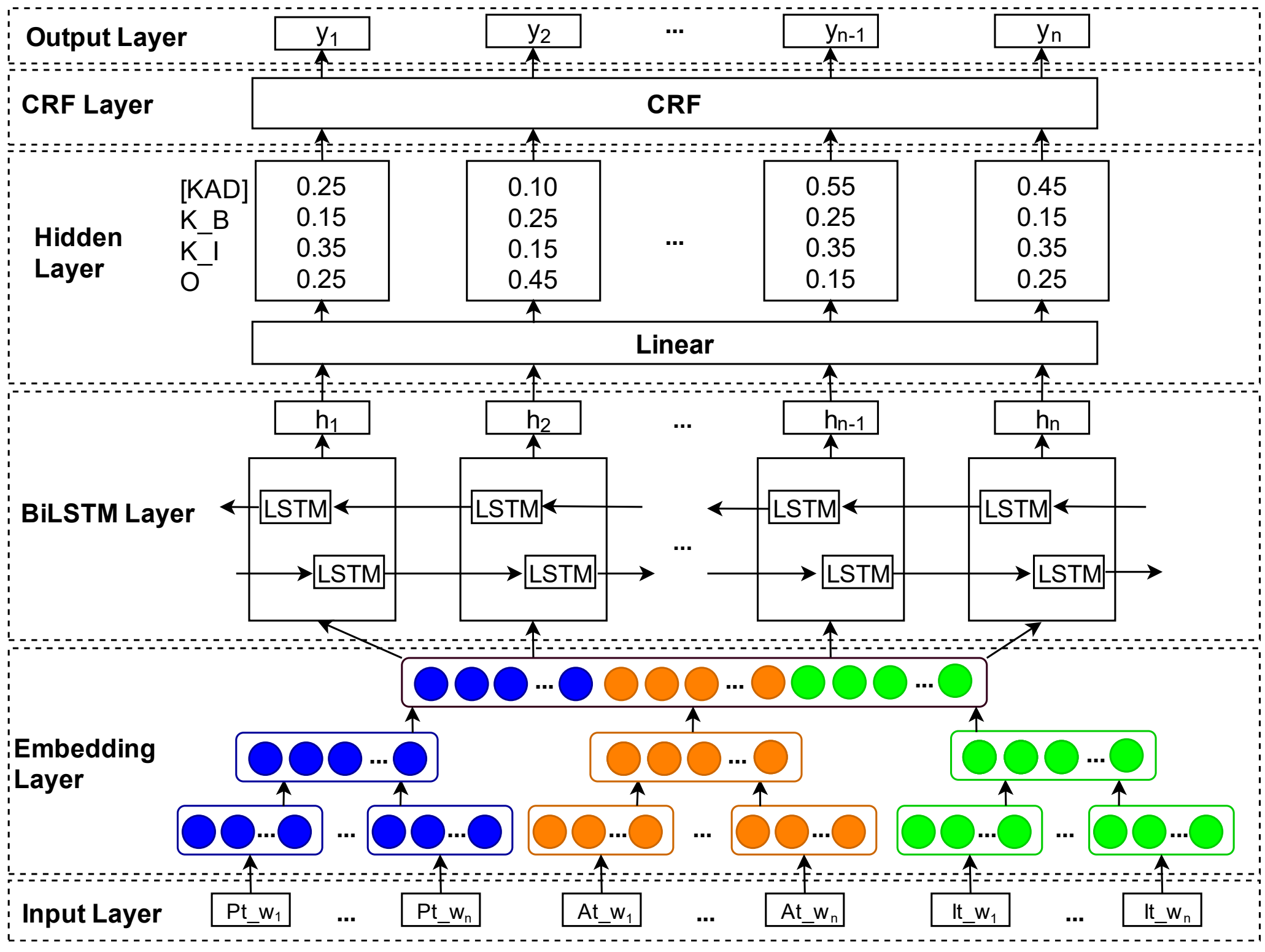

***Note*:** (1) Pt_w: Paper text words, At_w: Audio text words, It_w: Image text words.
(2) Blue: Vectorized representation of the text in the academic paper, Orange: Vectorized representation of the text in the audio, Green: Vectorized representation of the text in the image.

**Figure 4. BiLSTM-CRF Model Architecture Diagram**

### *Evaluation Method for Keyword Extraction*

In the evaluation of keyword extraction effectiveness, a precision matching method is used to compare the predicted keyword list generated by the model with the actual keyword list. This paper uses the keywords provided by the authors as the gold standard. The number of predicted keywords that exactly match the actual keywords is calculated. This paper employs the $F_1$ score ($F_1$) to evaluate the effectiveness of keyword extraction, with the specific calculation method shown in equations (1)-(3):

$$P = \frac{TP}{TP + FP} \times 100\% \quad (1)$$

$$R = \frac{TP}{TP + FN} \times 100\% \quad (2)$$

$$F_1 = 2 \times \frac{P \times R}{P + R} \times 100\% \quad (3)$$

Where TP represents the number of true positive samples predicted by the model. FP is the number of false positive samples predicted by the model. FN is the number of false negative samples predicted by the model.

### Experimental Setup

In this study, experiments are conducted using text, image and audio data. For the TextRank algorithm, the word co-occurrence window is set to 3, the maximum number of iterations is 100, and the convergence threshold is 1e-4. SVM and CRF models are randomly split into training and testing sets in a 7:3 ratio. In the SVM algorithm, due to the significant difference in the number of keywords and non-keywords, under sampling is applied to balance the number of positive and negative samples. The CRF model considered both word features and combination features. The BiLSTM-CRF and BERT-BiLSTM-CRF models are randomly split into training, validation, and testing sets in an 8:1:1 ratio. For the BiLSTM-CRF model, the dimensions of the embedding layer and hidden layer are 256, the learning rate is 1e-3, and the number of epochs is 20. For the BERT-BiLSTM-CRF model, the dimension of the hidden layer is set to 256, the learning rate is 2e-5, and the number of epochs is 20.

### Experimental Results Analysis

The experimental results of multimodal keyword extraction are presented in Table 2. The table demonstrates the performance of different keyword extraction models under various evaluation metrics ($F_1$@3, $F_1$@5, $F_1$@10) when using different types of text (paper, audio, image and concatenation of all three types of text). Under the condition of $F_1$@K, the bolded $F_1$ value indicates that the model performs the best on the corresponding modality of text.

| Model | $F_1$@3 | | | | $F_1$@5 | | | | $F_1$@10 | | | |
|---|---|---|---|---|---|---|---|---|---|---|---|---|
| | **Pt** | **At** | **It** | **Mt** | **Pt** | **At** | **It** | **Mt** | **Pt** | **At** | **It** | **Mt** |
| TF-IDF | **8.16** | 7.80 | 5.81 | 7.95 | 8.50 | **8.60** | 6.34 | 8.18 | 7.60 | **8.09** | 6.02 | 7.52 |
| TextRank | **5.89** | 4.84 | 5.83 | 5.26 | **6.69** | 5.65 | 6.52 | 6.38 | **6.67** | 6.12 | 6.36 | 6.39 |
| SVM | **10.93** | 9.67 | 6.82 | 10.72 | 13.14 | 11.64 | 8.65 | **13.23** | 14.25 | 12.47 | 10.26 | **15.40** |
| CRF | **16.27** | 13.21 | 10.77 | 15.72 | **17.17** | 14.01 | 11.01 | 16.85 | **16.98** | 14.64 | 11.02 | 16.56 |
| BiLSTM-CRF | 13.20 | 9.71 | 8.21 | **13.94** | 13.29 | 10.35 | 8.19 | **14.62** | 12.46 | 9.56 | 8.39 | **13.69** |
| Bert-BiLSTM-CRF | 14.99 | 11.87 | 13.88 | **15.34** | 15.74 | 12.62 | 13.88 | **16.22** | **15.96** | 12.27 | 12.28 | 15.61 |

***Note***: Pt: Paper text, At: Audio text, It: Image text, Mt: Concatenation of paper, audio and image texts

**Table 2. The Experimental Results of Multimodal Keyword Extraction**

From Table 2, it can be observed that the CRF model performs the best in extracting keywords from text, followed by the BERT-BiLSTM-CRF and BiLSTM-CRF models, while the performance of the SVM model is slightly inferior. In the TF-IDF model, when the standard number of keywords is set to 5 and 10, its performance in extracting keywords from audio text is better than the other three types of text; when set to 3, its performance is best in paper text. The TextRank model consistently outperforms other text types in keyword extraction from paper text. In the SVM model, when the standard number of keywords is set to 5 and 10, its performance in keyword extraction from concatenation of all three text types is better than the other three text types; when set to 3, its performance is best in paper text. In the CRF model, the keyword extraction performance from paper text is consistently better than other text types. In the deep learning model BiLSTM-CRF, its performance in keyword extraction from concatenation of all three text vectors is better than other text types. In the BERT-BiLSTM-CRF model, when the standard number of keywords is set to 3 and 5, its performance in keyword extraction from concatenation of all three text vectors is better than other text types; when set to 10, its performance is best in paper text. The poor performance of image text in keyword extraction task may be attributed to noise such as background affecting the accuracy of Optical Character Recognition (OCR) text.

Image text performs the worst in keyword extraction task compared to the other three types of text. This is attributed to noise such as background interference affecting Optical Character Recognition (OCR) text, resulting in either the inability to recognize text in images or inaccurate recognition of text. Poor-quality image text leads to subpar performance of both unsupervised and supervised keyword extraction models. While there is some improvement in the performance of keyword extraction models on audio text, it still does not achieve optimal results. This is due to factors such as background noise and the quality of recording devices affecting audio text quality, thereby impacting the accuracy of keyword extraction. Among the models, SVM, BiLSTM-CRF and BERT-BiLSTM-CRF exhibit better performance. SVM's robust generalization and ability to handle high-dimensional data effectively utilize feature information from various types of text, mapping different features into high-dimensional space and finding hyperplanes to separate keywords from non-keywords. BiLSTM-CRF and BERT-BiLSTM-CRF leverage advantages in sequence data and language modeling, learning semantic and contextual information within text sequences to integrate features from different modalities comprehensively, considering the correlated information among different modality texts, thus enhancing model performance. Compared to shorter audio and image texts, medium-length paper texts typically contain richer information, avoiding the issues of either sparse information due to overly lengthy text or missing information due to excessively short text. Therefore, medium-length paper texts demonstrate outstanding performance in SVM, CRF, BiLSTM-CRF and BERT-BiLSTM-CRF models.

## Case Study

We randomly select an instance from the test set to intuitively illustrate the impact of different text modalities and text concatenation on keyword extraction. The top ten predicted keywords from the keyword extraction model are selected and matched exactly with the standard keywords. In supervised learning models, when the number of extracted keywords is less than 10, all keywords are given. The table presents the paper title and the keywords provided by the authors, which serve as the gold standard. Detailed content is shown in Table 3. Regardless of whether it's a supervised or unsupervised keyword extraction method, performance on paper text consistently outperforms that on audio and image text. However, keywords that cannot be extracted from paper text may be found in audio or image text. For example, in the BiLSTM-CRF model, keyword like 'deterministic polishing' is extracted from audio text but not present in the paper text results, demonstrating the complementary nature of texts from different modalities in terms of information. Except for the TF-IDF model, extraction results of other models on concatenated text are equal to or better than those on individual paper, audio and image text. This indicates that using text concatenation methods can enhance the model's performance in keyword extraction.

<table>
<tr><td colspan="5">Title of paper: Deterministic Polishing of Replicating Grazing-Incidence Mandrels (https://doi.org/10.1117/12.2532198)</td></tr>
<tr><td colspan="5">Keywords provided by the authors: deterministic polishing, grazing-incidence, x-ray, optics</td></tr>
<tr><th>Modality / Model</th><th>Paper text</th><th>Audio text</th><th>Image text</th><th>Multimodal text</th></tr>
<tr><td>TF-IDF</td><td>mandrel,<br>polishing,<br>mandrel feed-rate,<br>deterministic polishing,<br>grazing-incidence mandrel,<br>magixs mandrel,<br>replicating mandrel,<br>dimensional polishing,<br>mandrel error,<br>zeeko polishing.</td><td>mandrel,<br>magixs mandrel,<br>wear,<br>entire mandrel,<br>polishing,<br>bonnet,<br>cgh,<br>wear pattern,<br>dimensional wear,<br>shell.</td><td>mandrel,<br>metrology,<br>mandrel shell,<br>cgh,<br>grazing-incidence mandrel,<br>magixs mandrel,<br>mandrel shell separation,<br>magixs mandrel richardson,<br>shell,<br>deconvolution metrology.</td><td>mandrel,<br>polishing,<br>mandrel feed-rate,<br>mandrel shell,<br>magixs mandrel,<br>grazing-incidence mandrel,<br>replicating mandrel,<br>mandrel error,<br>mandrel surface,<br>figured mandrel.</td></tr>
<tr><td>TextRank</td><td>polishing,<br>mandrel,<br>deterministic polishing,<br>polishing parameters,<br>polishing processes,<br>polishing data,<br>zeeko polishing,<br>polished section,<br>polishing run,<br>single polishing.</td><td>error,<br>polishing,<br>mandrel,<br>right,<br>space,<br>alignment error,<br>pattern,<br>wear pattern,<br>wear,<br>point spacing.</td><td>metrology,<br>mandrel,<br>mandrel shell,<br>cgh,<br>shell,<br>magixs mandrel richardson lucky deconvolution metrology,<br>theoretical residual error,<br>results conte initial metrology,<br>deterministic polishing,<br>mandrel shell separation.</td><td>polishing,<br>mandrel,<br>deterministic polishing,<br>polishing parameters,<br>polishing data,<br>polishing processes,<br>mandrel error,<br>polishing shows,<br>zeeko polishing,<br>polished section.</td></tr>
<tr><td>SVM</td><td>deterministic polishing,<br>mandrel,<br>polishing,<br>polishing parameters,<br>deterministic,<br>wear,<br>wear pattern,<br>grazing-incidence optics,<br>metrology,<br>grazing-incidence.</td><td>mandrel,<br>second half power diameter,<br>wear pattern,<br>mandrel,<br>half power diameter,<br>arc second half power,<br>polishing,<br>second half power,<br>dimensional fit,<br>arc second half power diameter.</td><td>metrology,<br>mandrel,<br>deterministic polishing,<br>deterministic,<br>time ultrahigh resolution metrology,<br>time ultrahigh resolution metrology data,<br>shell,<br>cgh,<br>dimensional,<br>time ultrahigh resolution.</td><td>mandrel,<br>deterministic polishing,<br>polishing,<br>wear,<br>deterministic,<br>grazing-incidence mandrel,<br>richardson lucy deconvolution,<br>wear pattern,<br>richardson lucy,<br>metrology.</td></tr>
<tr><td>CRF</td><td>x-ray,</td><td>cgh,</td><td>cgh,</td><td>deterministic polishing,</td></tr>
</table>

| | | | | |
|---|---|---|---|---|
| | replicated optics, deterministic polishing, x-ray optics, mandrel polishing, replicating mandrel, polishing, replicating, x-ray spectrometer, cgh. | deterministic polishing, wear. | deterministic polishing. | surface metrology, x-ray, replicated optics, x-ray optics, replicating, polishing. |
| BiLSTM-CRF | x-ray optics, grazing-incidence x-ray optics, grazing-incidence optics, optics, interferometric metrology, cgh, x-ray spectrometer, wolteri, grazing-incidence mirrors, x-ray. | cgh, polishing, deterministic polishing, national ignition, taylor series. | ultrahigh resolution, x-ray. | grazing-incidence, calibrated, replicating, wolteri, x-ray, cgh, artxc, polishing, grazing-incidence optics, optics. |
| BERT-BiLSTM-CRF | grazing-incidence optics, grazing-incidence x-ray, grazing-incidence x-ray optics, x-ray optics, grazing-incidence, x-ray, deterministic polishing, metrology, optics, deterministic. | deterministic polishing, calibrate, telescope, national ignition facility, deconvolution, cgh, alignment, polishing, grazing-incidence optics, wear pattern. | deterministic polishing, polishing, series. | x-ray spectrometer, national ignition facility, deterministic polishing, calibrate, grazing-incidence x-ray, x-ray, grazing-incidence optics, grazing-incidence, interferometry, grazing-incidence x-ray optics. |

***Note***: Green mark represents correctly predicted keywords, blue mark represents partially predicted words for the target answers.

**Table 3. The Results of Different Keyword Extraction Models based on Different Modalities (@K = 10).**

## CONCLUSION AND FUTURE WORKS

The objectives of this study are to build a multimodal dataset of academic paper for keyword extraction task, investigate the performance of different modalities of text information in keyword extraction task and explore the impact of text fusion from multimodal data on keyword extraction performance. The results reveal that text modalities from paper, image and audio each exhibits distinct characteristics. Among them, textual data from paper performs the best in keyword extraction task, followed by audio text with slightly lower performance and image text performs the worst. Traditional models such as SVM and CRF demonstrate effective keyword extraction performance when dealing with different modalities of text. On the other hand, deep learning models such as BiLSTM-CRF and BERT-BiLSTM-CRF achieve promising results when concatenating multimodal text vectors, suggesting that fusion of multimodal information contributes to enhanced keyword extraction performance. These findings underscore the potential and importance of multimodal information fusion in keyword extraction task.

For the newly constructed multimodal dataset of academic paper, the experiments only utilize the textual content from paper, image and audio, without considering the unique features of each modality. For instance, in slide images, specific font styles, layout characteristics, and visually emphasized text elements (e.g. larger font sizes, bold fonts and distinct colors) may indicate keywords. Audio features include speech energy, spectral centroid, and the duration of time spent by speakers on presenting images which is another crucial feature as important content tends to be elaborated upon for a longer duration. Additionally, there exist correlations between the three modalities, and harnessing this correlated information could potentially enhance the effectiveness of keyword extraction. Finally, due to the powerful language comprehension, self-learning, and generalization capabilities of large language models, they may achieve better performance in multi-modal keyword extraction task.

## ACKNOWLEDGMENTS

This paper was supported by the National Natural Science Foundation of China (Grant No. 72074113).